# USING PROGRAMMABLE DRONE IN EDUCATIONAL PROJECTS AND COMPETITIONS

**P. Petrovič, P. Verčimák**

*Comenius University Bratislava (SLOVAKIA)*

## Abstract

The mainstream of educational robotics platforms orbits the various versions of versatile robotics sets and kits, while interesting outliers add new opportunities and extend the possible learning situations. Examples of such are reconfigurable robots, rolling sphere robots, humanoids, swimming, or underwater robots. Another kind within this category are flying drones. While remotely controlled drones were a very attractive target for hobby model makers for quite a long time already, they were seldom used in educational scenarios as robots that are programmed by children to perform various simple tasks. A milestone was reached with the introduction of the educational drone Tello, which can be programmed even in Scratch, or some general-purpose languages such as Node.js or Python. The programs can even have access to the robot sensors that are used by the underlying layers of the controller. In addition, they have the option to acquire images from the drone camera and perform actions based on processing the frames applying computer vision algorithms. We have been using this drone in an educational robotics competition for three years without camera, and after our students have developed several successful projects that utilized a camera, we prepared a new competition challenge that requires the use of the camera. In the article, we summarize related efforts and our experiences with educational drones, and their use in the student projects and competition.

## 1 INTRODUCTION

Robot competitions are an important factor in propelling the development of sensors, actuators, and other technology, software frameworks, algorithms, and ideas and in motivating primarily young people to be more interested in technology, to learn more about controlling, programming, and building robots in depth and breadth, and to gently steer their interest towards the areas of science and technology. For the reasons of availability and accessibility, most of the robot competitions – whether they are organized more in an educational or more in hobby context – utilize either popular robotic sets, or some sort of ground vehicles – smaller or larger car-like or trolley-like robots performing all kinds of navigational, or manipulation tasks with various constraints and tasks. Flying robots (unmanned aerial vehicles), except of a few professional events such as Mohamed Bin Zayed International Robotics Challenge [1, 2], were traditionally a domain of a separate model maker communities, and they were associated with high purchase and maintenance costs, required a lot of expertise and tinkering, and they were typically used in outdoor environments. See figure 1 for an example of popular ZMR 250 Mini DIY Quadcopter. After a prevailing demand, we have built one such robot in our elementary school robot club, followed by 11 more in an electronics summer camp in 2017, each in a total price of about 120 Eur including remote controller and 3-cell LiPo battery. It is controlled by the then trendy open-source CC3D LibrePilot controller.

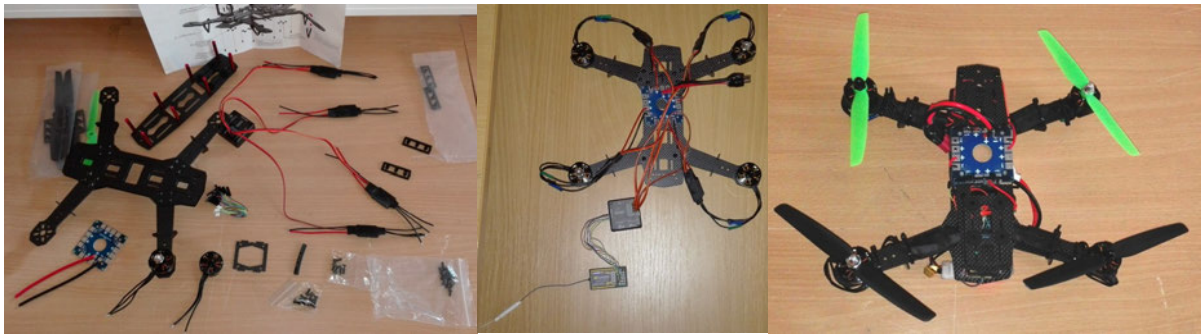

*Figure 1. Building the ZMR250 mini diy quadcopter in our robot club.*

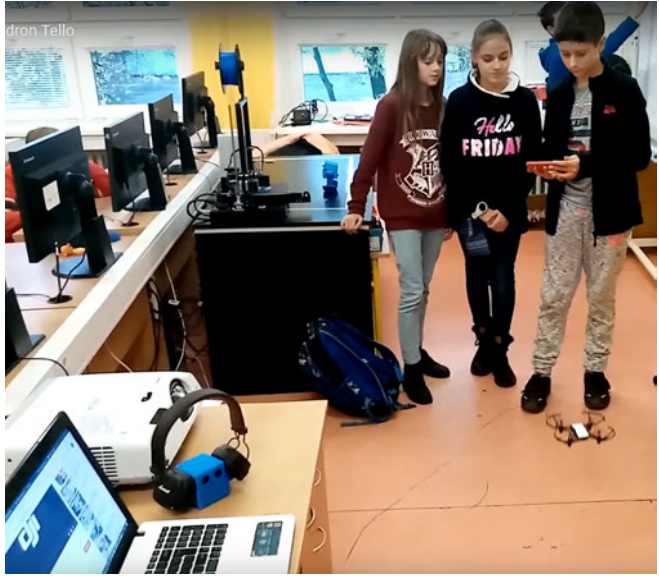

*Figure 2. Members of our robot club exploring the new Tello robot.*

Most of such drones were remotely controlled by a human over wireless radio link. Their stability was fragile due to noisy sensors and driving such drones required skills and practice. Almost four years ago, we have purchased a lightweight programmable drone DJI Ryze Tello [3, 4] for the price of approximately 100 Eur with unmatched stabilization quality and ease of control: since then, this robot has been used (and crashed) in many occasions, and with very few exceptions, any child or adult person who has attempted to remotely steer this robot from an Android phone over wifi connection became a successful pilot after just a one minute of instructions, see figure 2. This anecdotic evidence suggests that the times have changed, and programmable flying robots are entering the segment of educational robotics, and thus will also be more present at robot competitions.

## 2 RELATED WORK

Let us first discuss a few examples of previous or existing robot competitions with drone categories.

### 2.1 PRIA Aerial Showcase

Practical Robotics Institute Austria in cooperation with HTL tgm, the largest technical college in Vienna has been organizing the European Conference on Educational Robotics (ECER) [5], an international scientific conference for students, for more than 10 years. One of the main parts of this event is the European round of the Botball robot competition, globally organized by KISS Institute for Practical Robotics (KIPR) in Oklahoma. Botball [6] is "*a team-oriented robotics competition, where each team receives a defined robotics set consisting of metal and LEGO parts, sensors and actuators, two controllers and a standard vacuum cleaner robot. The competition task is redesigned each year and is not released until the two-day workshop at the beginning of the Botball season.*" Some other challenges take place at the same time, such as PRIA Underwater (organized in 4 pre-covid years), or PRIA Aerial, now called PRIA Aerial Showcase, which is a European version of KIPR's Aerial. KIPR also offers Aerial Botball Challenge Program [7], which is a "curriculum-based program for upper elementary, middle, and high school students that teaches computer science skills and the Python programming language in a team setting through hands-on use of a Tello EDU minidrone" that culminates at Aerial Challenge Days, competition events where student teams work to solve progressively more difficult challenges. For instance, the PRIA Aerial challenge in 2020 (and remains the same for 2023) was a search and rescue mission, where the drone is to bring objects of different types (radio masts and rescue kits, red or green cubes of 5x5 cm) to randomly placed areas that must be identified and located first, see figure 3. The mountain area requires the drone to fly by high boxes, where the opening is marked with green and red borders. Red cubes can be attached to the drone at the start, while the green ones must be picked up by the drone itself. Teams may place some additional guiding markers (such as light) in certain location on the field. Scoring in the challenge is accumulative for partial achievements, such as for moving each of the cubes out of the starting area to the base area (+1), botguy area (+10), or botguy zone (+30), landing within various areas (+20/30/50), and more.

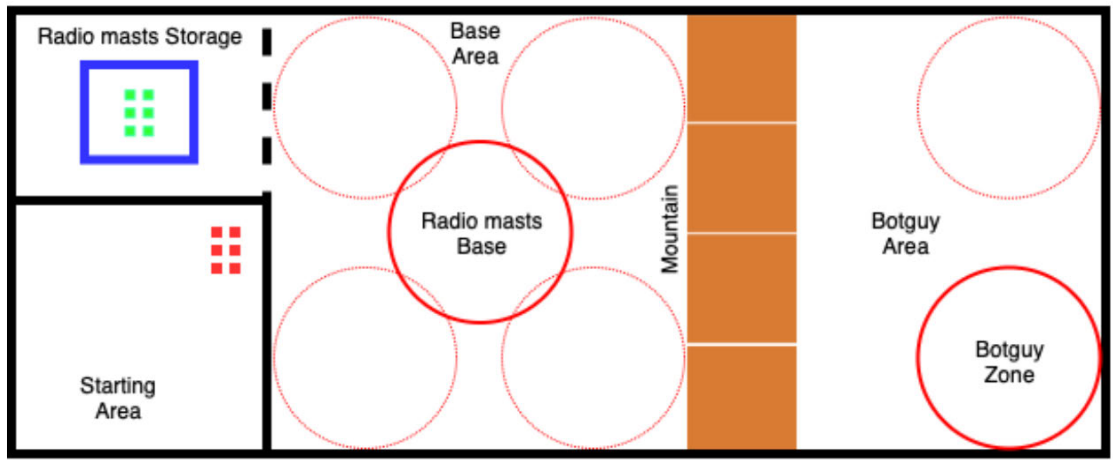

*Figure 3. PRIA Aerial challenge 2020-2023, dimensions of the field are 600x245 cm [7].*

Participants are obliged to only use approved drone models. For the 2023 season, these are Parrot AR Drone 2.0, Parrot Bebop 1, Parrot Bebop 2, Parrot Mambo, Parrot Anafi, Intel Aero RTF, DJI Spark, RYZE DJI Tello, and Coex Clover Drone.

## 2.2   MathWorks Minidrone Competition

Last year MathWorks announced another round of drone competition [8], which they organize at different regions around the World and which has two stages – a qualification simulated round, and on-site live round, which again can be in-person (4 events) or virtual (1 event). Interesting open-source solutions to this challenge already exist, for instance [9, 10]. Teams of 2-4 persons submit their solutions on-line for evaluation, and the best qualifying teams evaluated by the MathWorks engineers will participate at live rounds. Most of the events are open to students, while the Japan round is open to industry and academia, and the virtual EMEA event is open for anyone from that region. Participants must use Matlab/Simulink software to implement one part of the control algorithm as prepared by the MathWorks engineers, however each participating team will receive a complimentary license. The simulation rounds use a dedicated Simulink module, and the algorithms advancing to the live rounds run on Parrot Mambo minidrone. Other instances of this competition exist [11]. The task in this challenge is to follow a 10cm wide piecewise linear colored line (the color will be announced just before the competition round) containing no curves, and land at a 20cm circle that follows immediately the end of that line. The scores are based on the accuracy of the line following and landing and in case of successful landing also on the time used.

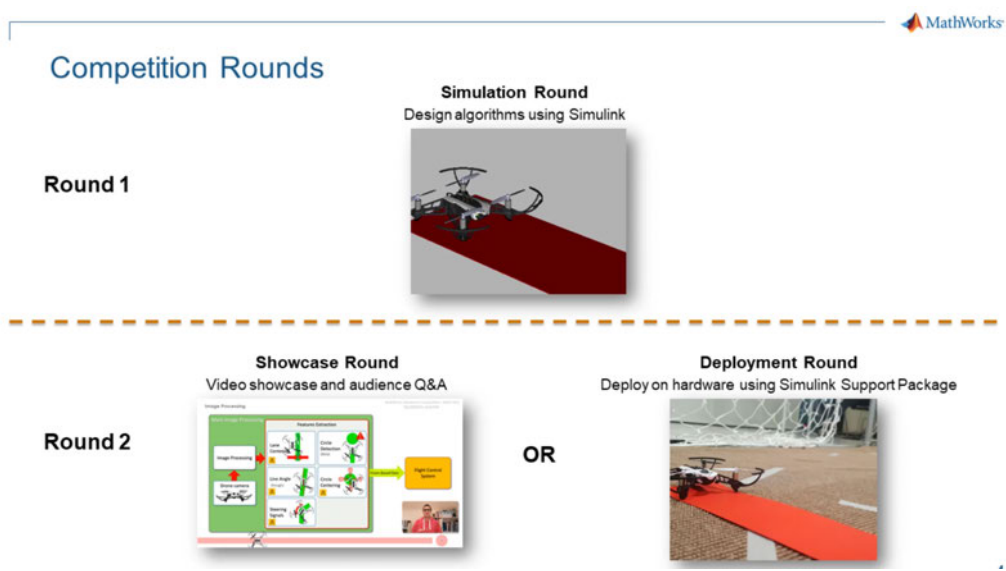

*Figure 4. MathWorks Minidrone competition outline [8].*

## 2.3 Istrobot Flying Challenge

Istrobot is a traditional robot contest in Slovakia (still) organized since the year 2000 [12]. The main organizer is the association Robotika.SK, which one of the authors is a member of. Among others, its most popular category is the line-follower, enriched with various selection of obstacle types (such as tunnel, oil spill, bridge, brick obstacle, curtain, interrupted lines, loops, changing the line color), and it attracts mostly beginners, but sometimes even seasoned skilled roboticists. A separate "free-ride" category is an exhibit, where teams from various robot clubs or technical schools bring any robotic designs and display them for the visitors as the event is not only meant for the participants, but also as a dissemination event towards public. Istrobot used to include a traditional micro-mouse maze solving challenge, and since the year 2012 it invented its own Ketchup house challenge, where two robots compete in acquiring and transporting tomato cans. In the years 2015-2018, Istrobot was also a platform for its "Flying challenge" category. The goal was to build a flying machine that can complete a specified path with obstacles in a limited space and bring a useful cargo to the goal in the shortest possible time.

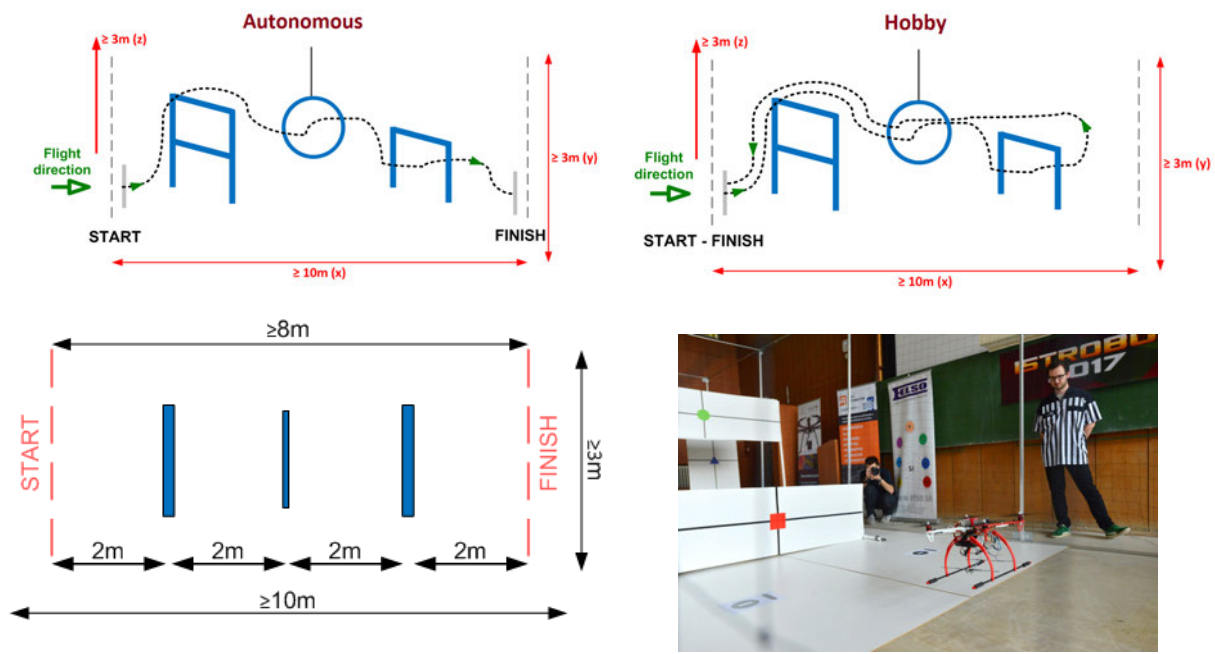

*Figure 5. Istrobot flying challenge.*

Participants were split into two groups: remotely controlled drones, and autonomous drones. Unfortunately, the category was completely dominated by the remotely controlled drones and therefore this category was dropped from the contest, just about the time of the appearance of those kinds of drones that are suitable for this type of indoor challenges.

## 3 PROGRAMMABLE DRONE DJI TELLO

Soon after the lightweight, low-cost, but high-quality stabilization drone Tello has been released, an add-on for the popular Scratch programming language was implemented [13]. Tello is normally remote-controlled from a mobile phone app with two on-screen joysticks and additional gestures. It can fly in all 4 directions, turn at a spot, perform flips in 8 directions, take off and land from/to a hand. Mobile phone control app is showing a 720p video stream captured by the drone's front-facing camera with built-in stabilization and transmitted over wifi TCP connection to the phone, where in-flight pictures can be taken, or videos recorded. The Scratch add-on is based on a wifi connection sending UDP control packets to the drone from a node.js program running in the background on the same computer as Scratch, communicating with Scratch add-on through local TCP sockets. See table 1 for a list of self-explanatory UDP packets recognized by Tello SDK [14]. This allowed children to augment its interactive Scratch projects with commands that controlled the flying drone, see figure 6.

5627

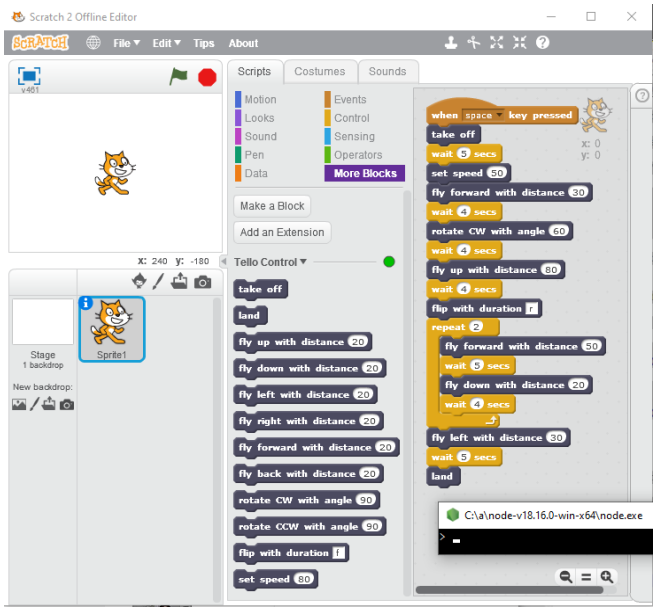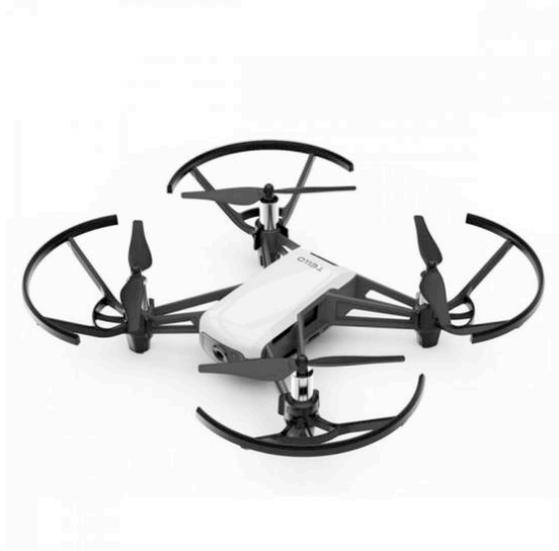

*Figure 6. Programming Tello in Scratch programming language (left) and the RYZE Tello drone (right).*

The high-quality stabilization is achieved thanks to the visual positioning system (VPS) mounted on the underside of the drone as well as time-of-flight (TOF) sensor and the inertial measurement unit (IMU). A disadvantage of VPS is that it fails to work properly on flat monochromatic or shining surfaces, and it is disturbed by moving objects, or when it is not in a proper height. In those situations, Tello autonomously starts finding a different spot, and this is outside the control of the user or the remote-control program. The propellers are well-protected by replaceable flexible shields, and the drone is capable of detecting collisions, in which case it cancels all the thrust and falls to the floor. Due to the (sad) popularity of the Python programming language, it did not take long for a library for controlling of Tello and obtaining the sensory information and video stream from its front-facing camera to appear. The DjiTelloPy is a Python open-source library allowing to emit commands supported by Tello SDK and read the response values as well as retrieve the video stream [15].

*Table 1. UDP Commands in Tello SDK ver.1.3 supported by original version of Tello.*

| *control:* | *fly:* | *turn:* | *set value:* | *read value:* | | |
|---|---|---|---|---|---|---|
| takeoff | up cm | cw deg | speed cm/s | speed? -> cm/s | baro? | -> m |
| land | down cm | ccw deg | rc lr fb ud yaw | battery? -> % | acceleration? | -> a b c |
| streamon | left cm | flip dir | wifi ssid pass | time? -> s | tof? | -> cm |
| streamoff | right cm | go x y z cm/s | | height? -> cm | wifi? | -> snr |
| emergency | forward cm | curve x1 y1 z1 x2 y2 z2 speed | | temp? -> degC | | |
| | back cm | | | attitude? -> p r y | | |

## 4 ROBOCUP JUNIOR IN SLOVAKIA

Since the first LEGO robotic sets (LEGO Dacta Control Lab) appeared in the schools in Slovakia and Czech Republic in the 90s, students from elementary and secondary schools annually met at national creative competitions [16]. With the introduction of RoboCup Junior initiative [17], we have extended the original competition in 2002 with RoboCup Junior categories – Soccer, Rescue, and Dance (OnStage), and later also simulated categories (CoSpace, Simulated Soccer, Erebus). A whole generation of in total about 2000 students benefited from the program since then, and many dozens of them could grow through participation in international rounds. In the fall of 2019 one of our regional organizers received feedback from one of the schools that they have a new programmable drone,and look for an event in which they could participate, and thus we have added a new category in 2020: Drone.

## 5  FLYING DRONE AT ROBOCUP JUNIOR IN SLOVAKIA 2020-2022

At that time, we could not find much software support for retrieving and processing the view from the drone's camera or the sensors, so our first mission [18] required to program the drone for a specified path, navigate the obstacles, reach target 70cm high table, pick up a victim that is waiting there to be rescued, and navigate the same trajectory back to the start in attempt for a successful landing. Figure 7 shows the racing track that we designed somewhat inspired by the abandoned Istrobot Flying Challenge: view from a perspective, front view, and side view. The rings had a diameter 1m, one touching the floor, the other being 1m above the ground. Drone could have been successfully programmed to follow this trajectory. The first year no participants were ready (the event was cancelled due to covid), in 2021 we had one participant (it was an on-line event), see figure 8, and in the third year we had about 5 teams participating.

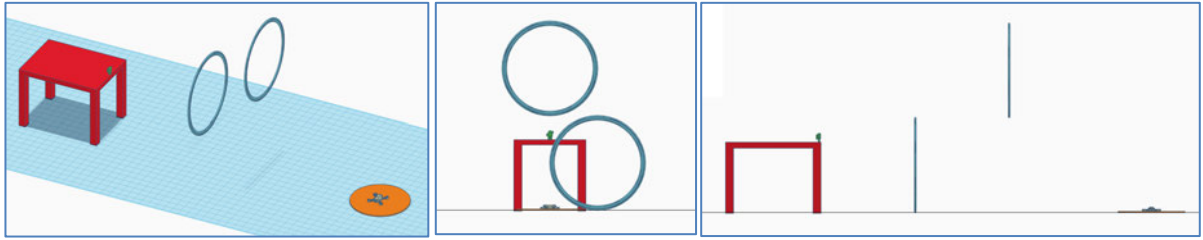

Figure 7. Layout for a drone competition in first three years, views from different angles.

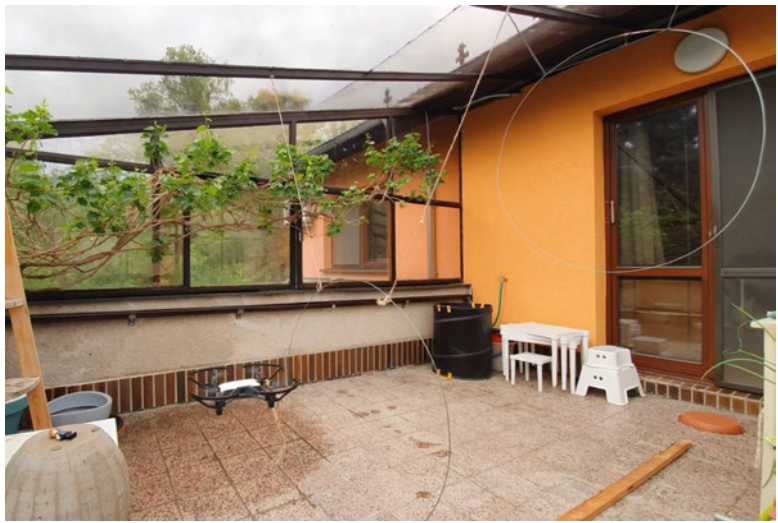

Figure 8. Robot in the middle of the mission, online participant in 2021.

The victim could have been any small object that the teams designed themselves. There were no limitations on how the object could be captured, lifted, and transported. Teams came up with various creative ideas of how to pick up their victims: some tried to use magnets, others a wire bent in a shape of a hook, yet others used Velcro. It was close, but none of them managed to pick it up and successfully complete the transport. As we discussed above, the task was difficult due to the behavior of the drone that it is not very happy to "see" anything else than a floor with a nice texture underneath, otherwise, it likes not to obey the intentions of the remote control. At that time, we were not fully aware of the extent this feature renders the task difficult. However, some teams managed to navigate to the victim and back and land successfully. The scores were gradually added for partial success, 7x10 points for navigating each phase, touching the rings incurred a small penalty. As in other RoboCup categories, some points (30) were allocated for technical interview and documentation where the teams had to explain their solutions.

Meanwhile at our Faculty, we asked students to investigate the possibilities to retrieve and process the frames from the drone's camera in their final project for Practical Seminar in Robotics. The first project used OCR neural network translating hand-written signs into instructions for robot [19], and the second was a simple line following robot [20]. In this case, we have mounted a 45º titled mirror placed in front of the drone's camera to change its direction to be looking downward, see figure 9. It is based on a design from [21].

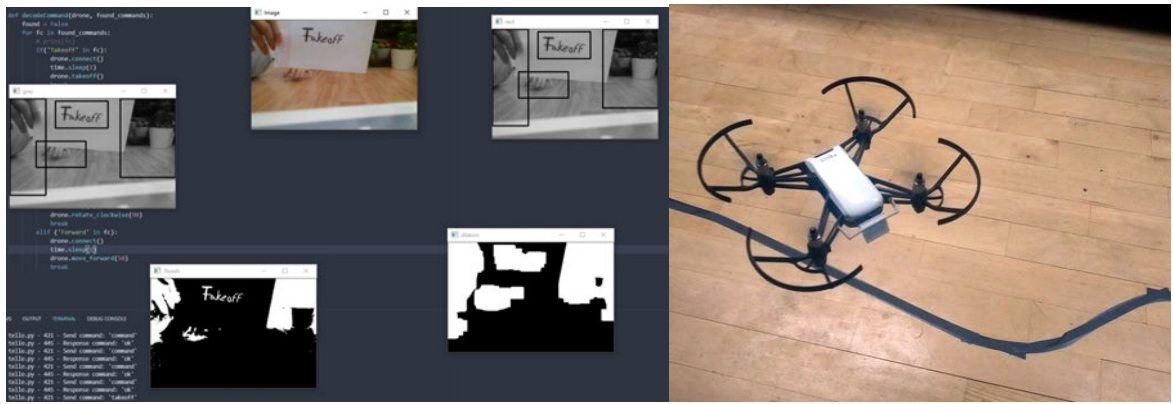

*Figure 9. Controlling drone with signs using optical character recognition (left), line-following robot (right).*

## 6  FLYING DRONE AT ROBOCUP JUNIOR IN SLOVAKIA 2023

After the two projects successfully confirmed feasibility of using camera of Tello drones for controlling their behavior, we opened a call for a final thesis in the undergraduate program of Applied Informatics at our faculty. One of the authors of this paper responded and selected that topic [22]. His task was to prepare a new challenge for 2023 that will require the use of vision, organize the challenge at our annual national event, evaluate it, and prepare tutorials for the secondary school students in Slovak language that will lower the entry-threshold to this contest. The tutorials should explain background theory, algorithms that could be useful, tools that are needed or helpful – how to install, how to use, provide code samples for a few solved challenges, and solve the challenge just to verify its difficulty and solvability.

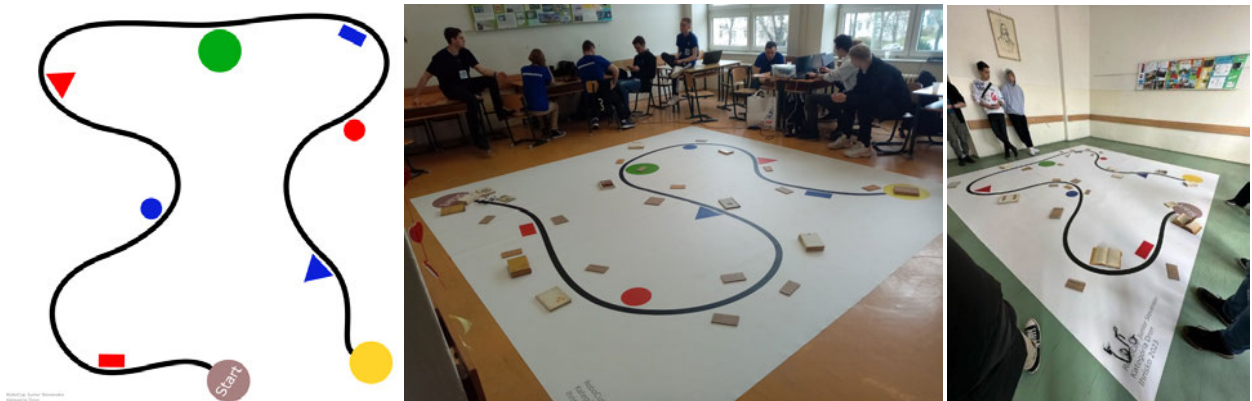

*Figure 10. Challenge 2023 – layout of the situation on the left – dimensions 4x4 m, workshop with the participants (with Peter Verčimák in top left corner of the image) - center, the actual contest on the right.*

The dates were somewhat restrictive, but the RoboCup Junior Slovakia was held at technical college in Prešov (SPŠE) on April 12[th] - 14[th] and despite of that, four teams have participated. The challenge was as follows [18] (see figure 10): Robot flies over white field partially covered with objects (here we used old books) to allow a correct use of VPS. It takes off at a designated location, follows 5cm black line, may pick a victim from a rescue pad in the middle of the field, continues following, and finally lands in the goal area. It should respond to colored shapes located nearby the line on any of its side:

- red rectangle (24 x 12 cm): drone will start video recording, and continues to follow the line
- red circle (d = 20 cm): drone must fly up to the height of at least 2m, continue so for at least 1m
- red triangle (a = 15 cm): drone rotates 360 degrees left, and continue
- blue rectangle (24 x 12 cm): drone stops recording video, saves it, and keeps following
- blue circle (d = 20 cm): drone must fly down to less than 1m for at least 1m while following

- blue triangle (a = 15 cm): drone rotates 360 degrees right, and continue
- green circle (d = 40 cm) contains a victim in the center, drone may pick it up, and then return to line to continue the following
- yellow circle (d = 40 cm) demarks the goal area.

The scoring is gradual, successful take-off initiates with 5 pts, requested flips/altitude changes gain 5 pts, every temporary leaving the line costs -5 pts, successfully following, and changing the heading of the drone facing the line direction earns 15-25 pts, recording the view as requested for 10 pts, 5-10 pts for landing depending on the precision, 10 pts for the victim rescue, and 30 pts for a technical interview and documentation. As before, victim can be anything the participants bring with the minimum dimensions of a little LEGO figure.

None of the teams earned maximum score, none was able to rescue a victim, but two were able to do some line-following and one was even able to complete about half of the course. More detailed evaluation showed that even though their school (a technical college in Bratislava) provided the students with several drones, and we have published the tutorials weeks before the contest, the participants failed to organize enough time before the contest to prepare and debug their code. We considered this year as a testing preliminary trial and attempt to help a better participation next year through early advertising, and communication with teams prior to the tournament. The author of the challenge organized a workshop for the participants, where they learned about the methods and algorithms, and tried some of those on a testing course track. He has also demonstrated a working complete solution without disclosing a full source code.

## 7   SUPPORT FOR THE CONTEST PARTICIPANTS

The author of the challenge has developed a tutorial for the participants from secondary schools that they can use to get more familiar with Tello programming and prepare for solving the challenge. It is published at github wiki [23], alongside the repository with the example code. The tutorial starts with instruction how to install OpenCV, Numpy, and DJITello libraries, explains and provides examples on using HSV color model for color detection, explains and shows examples on image thresholding, morphological operations, edge, shape and contour detection, video recording, and discusses the algorithmic ideas for drone line-following algorithms, and wraps everything to a library of functions that perform some elementary operations needed when solving the challenge. In this way, students get hands on several working examples with explanations, save a lot of time they would waste with the technical setup and dig right into the interesting part of the challenge, all they need is to offer several hours of studying and trying out the tutorial. After the contest, the author worked on providing new functionality for next year challenge: we already have person following, QR code and gesture recognition

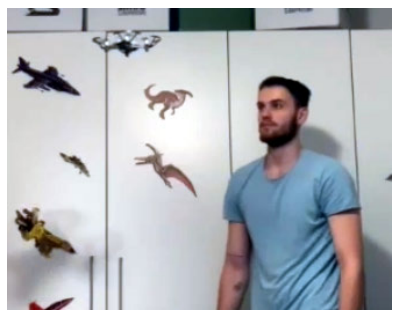

*Figure 11. Drone following a person.*

## 8   CONCLUSIONS

Unmanned aerial vehicles are a very promising technology with many civil applications capable of life-saving, and various other valuable support operations. Drones programming introduces novel situations and concepts that are not present with surface robotics: they navigate in 3D world, cope with a different type of inertial movement, are subject to a constant gravitational acceleration, process different sensory information, and may require different approach to control algorithms. We have demonstrated in this paper that the technology is mature and ready for wider use in educational process. We reviewed several past and existing flying drone competitions and showed the extrapolation of the technology to a more regular use in daily teaching and project work. A side product is an android application for the referees evaluating the robot challenge [22]. The main goal of this work is to take small steps in wider adoption and dissemination of ideas. For that end have organized three years of flying drone competition in Slovakia as part of our annual robot contest and prepared a tutorial for secondary school students that allows them to flatten the learning curve and overcome possible starting thresholds. We believe that flying drones in educational context have the potential to attract more students to science and technology, broaden understanding and acceptance of drones as useful tools in many applications, motivate students to study the topic of drones in more depth, and contribute to the further development of the technology. For the future, we plan to continue organizing the drone category, adding more challenges for both downward and forward-looking camera situations. The setup is a plausible motivation for the students to learn more about image processing algorithms thus unlocking an unlimited potential of very useful tasks and challenges for both completely autonomous missions as well as situations requiring human-robot interaction.

## ACKNOWLEDGEMENTS

The project was partially supported by Horizon Europe project TERAIS, GA 101079338.